\def\year{2021}\relax
\documentclass[letterpaper]{article} 
\usepackage{aaai21}  
\usepackage[switch]{lineno}  
\usepackage{times}  
\usepackage{helvet} 
\usepackage{courier}  
\usepackage[hyphens]{url}  
\usepackage{graphicx} 
\usepackage{natbib}  
\usepackage{caption} 
\usepackage{lipsum}

\usepackage[ruled,vlined,linesnumbered]{algorithm2e} 
\usepackage{booktabs} 

\usepackage{amsmath}
\usepackage{amssymb}
\usepackage{ifthen}
\usepackage{xcolor}
\newboolean{showcomments}
\setboolean{showcomments}{true}

\ifthenelse{\boolean{showcomments}}
  {\newcommand{\nb}[3]{
  {\color{#2}\small\fbox{\bfseries\sffamily\scriptsize#1}}
  {\color{#2}\sffamily\small$\triangleright~$\textit{\small #3}$~\triangleleft$}
  }
  }
  {\newcommand{\nb}[3]{}
  }

\usepackage{acronym}
\usepackage{xspace}
\acrodef{SOC}{sum of costs}
\acrodef{SIPP}{Safe interval path planning}
\acrodef{CBS}{Conflict-based search}
\acrodef{CCBS}{Continuous-time conflict-based search}
\acrodef{CCBS-DS}{Continuous-time conflict-based search with disjoint splitting}
\acrodef{ICBS}{Improved conflict-based search}
\acrodef{MAPF}{Multi-Agent Pathfinding}
\acrodef{ICTS}{Increasing Cost Tree Search}
\acrodef{MCCBS}{Multi-Constraint CBS}
\acrodef{CT}{Constraint Tree}
\acrodef{PC}{Prioritizing Conflicts}
\acrodef{DS}{Disjoint Splitting}

\newcommand{\cbs}{\ac{CBS}\xspace}

\newcommand{\ccbs}{\ac{CCBS}\xspace}
\newcommand{\ccbsds}{\ac{CCBS-DS}\xspace}

\newcommand{\cbsds}{{CBS-DS}\xspace}
\newcommand{\ct}{\ac{CT}\xspace}
\newcommand{\sipp}{\ac{SIPP}\xspace}
\newcommand{\soc}{\ac{SOC}\xspace}
\newcommand{\astar}{A$^*$\xspace}
\newcommand{\mapfr}{{MAPF}$_R$\xspace}
\newcommand{\mapf}{\ac{MAPF}\xspace}
\newcommand{\const}{\textit{constraints}\xspace}

\newcommand{\pc}{\ac{PC}\xspace}
\newcommand{\ds}{\ac{DS}\xspace}

\newcommand{\gsipp}{GSIPP\xspace}

\title{Improving Continuous-time Conflict Based Search*}
\author {

        Anton Andreychuk,\textsuperscript{\rm 1, 2}
        Konstantin Yakovlev, \textsuperscript{\rm 2, 3}
        Eli Boyarski, \textsuperscript{\rm 4} 
        Roni Stern \textsuperscript{\rm 4, 5}\\
}
\affiliations {
    \textsuperscript{\rm 1} Peoples' Friendship University of Russia (RUDN University)\\
    \textsuperscript{\rm 2} Federal Research Center for Computer Science and Control of Russian Academy of Sciences\\
    \textsuperscript{\rm 3} HSE University 
    \textsuperscript{\rm 4} Ben-Gurion University of the Negev \\
    \textsuperscript{\rm 5} Palo Alto Research Center \\
    
    andreychuk@mail.com, yakovlev@isa.ru, eli@boyar.ski,
    sternron@post.bgu.ac.il
}

\begin{document}

\maketitle


\begin{abstract}
Conflict-Based Search (CBS) is a powerful algorithmic framework for optimally solving classical multi-agent path finding (MAPF) problems, where time is discretized into the time steps. 
Continuous-time CBS (CCBS) is a recently proposed version of CBS that guarantees optimal solutions without the need to discretize time. 
However, the  scalability of CCBS is limited because it does not include any known improvements of CBS. 
In this paper, we begin to close this gap and explore how to adapt successful CBS improvements, namely, prioritizing conflicts (PC), disjoint splitting (DS), and high-level heuristics, to the continuous time setting of CCBS. 
These adaptions are not trivial, and require careful handling of different types of constraints, applying a generalized version of the \acf{SIPP} algorithm, and extending the notion of cardinal conflicts.
We evaluate the effect of the suggested enhancements by running experiments both on general graphs and $2^k$-neighborhood grids. 
CCBS with these improvements significantly outperforms vanilla CCBS, 
solving problems with almost twice as many agents in some cases and 
pushing the limits of multi-agent path finding in continuous-time domains.

\end{abstract}

\section{Introduction}

\acf{MAPF} is the problem of finding paths for $n$ agents in a graph such that 
each agent reaches its goal vertex and the agents do not collide with each other while moving along these paths. 
Many real-world applications require solving variants of \mapf, 
including managing aircraft-towing vehicles \cite{MorrisPLMMKK16}, video game characters \cite{Silver05}, office robots \cite{VelosoBCR15}, and warehouse robots \cite{WurmanDM07}. 
Solving \mapf optimally (for common objective functions) is NP-Hard~\cite{surynek2010optimization,yu2013structure}, 
but modern optimal \mapf algorithms can scale to problems with over a hundred agents~\cite{sharon2015conflict,BoyarskiFSSTBS15,felner2018adding,BCP,lazy-cbs,SurynekFSB16}.

However, such scaling was shown mostly on the classical version of the \mapf problem~\cite{stern2019multi}, which embodies several simplifying assumptions such as all actions have the same duration and time is discretized into time steps. \mapfr~\cite{walker2018extended} is a generalization of the classical \mapf problem in which actions' durations can be non-uniform, agents have geometric shapes that must be considered, and time is continuous. 
Handling continuous time is challenging because it implies an agent may wait in a location for an arbitrary amount of time, i.e., the number of \emph{wait actions} is infinite.  

Several recently proposed algorithms address the \mapfr problem or its variants, such as Extended ICTS (E-ICTS)~\cite{walker2018extended}, CBS with Continuous Time-steps (CBS-CT)~\cite{cohen2019optimal}, and 
\ccbs~\cite{andreychuk2019multi}. 
In this work, we propose several improvements to \ccbs that allow it to solve \mapfr problems with significantly more agents. 
\ccbs is based on the \cbs algorithm for classical \mapf, and the improvements we propose for \ccbs are based on known improvements of \cbs, namely \ds, \pc, and high-level heuristics. 
Adapting the \ds technique to the continuous-time setting of \mapfr requires solving a single-agent pathfinding problem with temporally-constrained action landmarks~\cite{karpas2009cost}. 
We show how to efficiently solve this pathfinding problem in our context by applying a generalized version of the \sipp algorithm~\cite{phillips2011sipp}. 
A naive applying of \pc to \ccbs is shown to be ineffective, and we propose an adapted version of \pc that can cut the number of expanded nodes significantly. 
The third \ccbs improvement we propose is an admissible heuristic function for \ccbs that require only a negligable amount of overhead when applied together with the \pc technique.

Finally, we evaluate the impact of these improvements individually and collectively on several benchmarks, including both roadmaps and grids. 
The results show that the number of \mapfr instances solved by \ccbs with all the proposed improvements compared to vanilla CCBS
has increased by 49.2\% --- from 3,792 to 5,659. 
In some cases, it can even solve problems with approximately twice the number of agents compared to vanilla CCBS and reduce the runtime up to two orders of magnitude.

\section{Background and Problem Statement}

In a \mapfr problem~\cite{walker2018extended}, the agents are confined to a weighted graph $G=(V, E)$ whose vertices ($V$) correspond to locations in some metric space, e.g. $\mathbb{R}^2$ in a Euclidean space, and edges ($E$) correspond to possible transitions between these location. 
Each agent $i$ is initially located at vertex $s_i\in V$ and aims to reach vertex $g_i\in V$. 
When at a vertex, an agent can either perform a \emph{move} action or a \emph{wait} action. 
A move action means moving the agent along an edge. 
We assume that the agent moves in a constant velocity and inertial effects are neglected. 
The duration of a move action is the weight of its respective edge. 
A wait action means the agents stays in its current location for some duration. 
The duration of a wait action can be any positive real value. Since we do not discretize time, the set of possible wait actions is uncountable. 

A \emph{timed action} is a pair $(a_i, t_i)$ representing that action $a_i$ (either move or wait) starts at time $t_i$. 
A \emph{plan} for an agent is a sequence of timed actions such that executing this sequence of timed actions moves the agent from its initial location to its goal location. 
The cost of a plan is the sum of the durations of its constituent actions. 
We assume that after finishing the plan the agent does not disappear but rather stays at the last vertex forever, but this ``dummy'' wait action does not add up to the cost of the plan.\footnote{This assumption is common in the MAPF literature.}

The plans of two agents are said to be \emph{conflict free} if the agents following them never collide, i.e. their \emph{shapes} never overlap. 
A \emph{joint plan} is a set of plans, one per each agent. 
A \emph{solution} to a \mapfr problem is joint plan whose constituent plans are pairwise conflict-free. 
The cost of a solution is its \soc, i.e., the sum of costs of its constituent plans.  
In this work, we are interested in solving \mapfr problems optimally, i.e., finding a  solution with a minimal cost. 
\ccbs~\cite{andreychuk2019multi} is a \cbs-based algorithm that does so. 
For completeness, we provide a brief description of \cbs and \ccbs below.

\subsection{\acf{CBS}}

\cbs~\cite{sharon2015conflict} is a complete and optimal algorithm for solving classical \mapf problems, i.e., \mapf problems where time is discretized and all actions have the same duration. 
\cbs works by finding plans for each agent separately, detecting \emph{conflicts} between these plans, and resolving them by replanning for the individual agents subject to specific \emph{constraints}.
A \cbs conflict in \cbs is defined by a tuple $(i,j,x,t)$ 
stating that agents $i$ and $j$ have a conflict in location $x$ (either a vertex or an edge) at time $t$. A \cbs constraint is defined by a tuple $(i,x,t)$, which states that agent $i$ cannot occupy $x$ at time $t$. 
To resolve a conflict $(i,j,x,t)$, \cbs replans for agent $i$ or $j$ or both, subject to \cbs constraints $(i,x,t)$ and $(j,x,t)$, respectively. 
To guarantee completeness and optimality, \cbs runs two search algorithms: a low-level search algorithm that finds paths for individual agents subject to a given set of constraints, and a high-level search algorithm that chooses which constraints to impose and which conflicts to resolve.

\paragraph{\cbs: Low-Level Search.}
In the basic \cbs implementation, the low-level search is a search in the state space of vertex-time pairs. 
Expanding a state $(v,t)$ generates states of the form $(v',t+1)$, where $v'$ is either equal to $v$, representing a wait action, or equal to one of the locations adjacent to $v$. 
States generated by actions that violate the given set of \cbs constraints, are pruned. 
\cbs runs \astar on this search space to return the lowest-cost path to the agent's goal that is consistent with the given set of \ac{CBS} constraints, as required.

\paragraph{\cbs: High-Level Search.}
The \cbs high-level search is a search in a binary tree called the \ct. 
In the \ct, each node $N$ represents a set of \cbs constraints $N.\const$
and a joint plan $N.\Pi$ that is consistent with these constraints. 
Generating a node $N$ involves settings its constraints $N.\const$ and running the low-level search to create $N.\Pi$. 
If $N.\Pi$ does not contain any conflict, then $N$ is a goal. 
Expanding a non-goal node $N$ involves choosing a conflict $(i,j,x,t)$ in $N.\Pi$ 
and generating two child nodes $N_i$ and $N_j$. 
Both nodes have the same set of constraints as $N$, plus a new \cbs constraint: $(i,x,t)$ for $N_i$ and $(j,x,t)$ for $N_j$. 
This type of node expansion is referred to as \emph{splitting node $N$ over conflict $(i,j,x,t)$}.
The high-level search finds a goal node by searching the \ct in a best-first manner, expanding in every iteration the \ct node $N$ with the lowest-cost joint plan.

\subsection{Continuous-Time Conflict Based Search (CCBS)}

To consider continuous time, \ccbs~\cite{andreychuk2019multi} reasons over the \emph{time intervals}, 
detects conflicts between \emph{timed actions}, 
and resolves conflicts by imposing constraints that specify the time intervals in which the conflicting timed actions can be moved to avoid the conflict.  
Formally, a \ccbs conflict is a tuple $(a_i, t_i, a_j, t_j)$, specifying that the timed action $(a_i, t_i)$ of agent $i$ has a conflict with the timed action $(a_j, t_j)$ of agent $j$.  
The \emph{unsafe interval} of timed action $(a_i, t_i)$ w.r.t. the timed action $(a_j, t_j)$, denoted $[t_i, t_i^u)$, is the maximal time interval starting from $t_i$ in which performing $a_i$ creates a conflict with performing $a_j$ at time $t_j$.
A \ccbs constraint is a tuple $(i, a_i, [t_i, t_i^u))$ specifying that agent $i$ cannot perform action $a_i$ in the time interval $[t_i, t_i^u)$. 
To resolve a \ccbs conflict, \ccbs generates two new \ct nodes, 
where it adds the constraint $(i, a_i, [t_i, t_i^u))$ to one node 
and the constraint $(j, a_j, [t_j, t_j^u))$ to the other. 

The low-level planner of \ccbs is an adaptation of the \sipp algorithm~\cite{phillips2011sipp}. 
\sipp was originally designed to find time-optimal paths for an agent moving among the dynamic obstacles with known trajectories. 
\sipp runs a heuristic search in the state-space of $(v, [t, t'])$ tuples, where $v$ is the graph vertex and $[t, t']$ is a \emph{safe interval} of $v$, i.e. a maximal contiguous time interval in which an agent can stay or arrive at $v$ without colliding with a moving obstacle. As numerous obstacles may pass through $v$ there can exist numerous search nodes corresponding to the same graph vertex but different time intervals in the \sipp search tree. 

The \ccbs low-level search is based on \sipp except for how it handles the given \ccbs constraints. 
Instead of dynamic obstacles, the low-level \ccbs computes safe intervals for each vertex $v$ with respect to the \ccbs constraints imposed over wait actions at $v$. 
Initially, vertex $v$ has a single safe interval $[0, \infty)$. 
Then, for every \ccbs constraint $(i, a_i, [t_i, t_i^u))$ 
where $a_i$ is a wait action at vertex $v$, we split the safe interval for $v$ 
to arriving before $t_i$ and to arriving after $t_i^u$. 
\ccbs constraints imposed over the move actions are integrated into the low-level search by modifying the constrained actions, as follows. 
Let $v$ and $v'$ be the source and target destinations of $a_i$. 
If the agent arrives to $v$ at $t\in [t_i, t^u_i)$ then we remove the action that moves it from $v$ to $v'$ at time $t$, and add an action that represents waiting at $v$ until $t^u_i$ and then moving to $v'$.

\section{Disjoint Splitting for \ccbs}

The first technique we migrate from \cbs to \ccbs is called Disjoint Splitting (\ds)~\cite{li2019disjoint}. 
\ds is a technique designed to ensure that expanding a \ct node $N$ creates a disjoint partition of the space of solutions that satisfy the constraints in $N.\const$.  
That is, every solution that satisfies $N.\const$ is in exactly one of its children. Observe that this is not the case in \cbs: for a conflict $(i,j,v,t)$ there may be solutions that satisfy both $(i,v,t)$ and $(j,v,t)$. 
This introduces an inefficiency in the high-level search.

To address this inefficiency, \cbs with \ds (\cbsds) introduces the notion of \emph{positive} and \emph{negative} constraints.   
A \emph{negative} constraint $\overline{(i, x, k)}$ is the regular \cbs constraint stating that agent $i$ \emph{must not} be at $x$ at time step $k$.
A \emph{positive} constraint $(i, x, k)$ means that agent $i$ \emph{must} be at $x$ at time step $k$. When splitting a \ct node $N$ over a \cbs conflict $(i, j, x, k)$, 
\cbsds chooses one of the conflicting agents, say $i$, 
and generates two child nodes, one with the negative constraint $\overline{(i, x, k)}$ 
and the other with the positive constraint $(i,x,k)$. 
Deciding on which agent, either $i$ or $j$ to split on, does not affect the theoretical properties of the algorithm, and several heuristics were proposed~\cite{li2019disjoint}.

The low-level search in \cbsds 
treats each positive constraint as a special type of \emph{fact landmark}~\cite{richter2008landmarks}, i.e., a fact that must be true in any plan.
The \cbsds low-level search generates a plan that satisfies these fact landmarks by 
planning to achieve these fact landmarks in ascending order of their time dimension. 
This effectively decomposes the low-level search to a sequence of simpler search tasks, searching for path one fact landmark to the next one. 
The agent's goal is set a the last fact landmark, to ensure the agent reaches it eventually.

\subsection{Positive and Negative Constraints in \ccbs}

A \ccbs constraint $(i, a_i, [t_i, t_i^u))$ can be stated formally as follows:
\[
\forall t\in [t_i, t_i^u): (a_i, t) ~~~ \text{is \emph{not} in a plan for agent $i$}
\]

This is a negative constraint from a \ds perspective. The corresponding positive constraint is therefore the inverse: 
\[
\exists t\in [t_i, t_i^u): (a_i, t) ~~~ \text{is in a plan for agent $i$}
\]
This mean that agent $i$ must perform $a_i$ at some moment of time from the given interval. Thus a positive constraint in \ccbs is an \emph{action landmark}~\cite{karpas2009cost}, i.e., the action that must be performed in any solution. Next, we show how the low level search of \ccbsds is able to find a plan that achieves all these action landmarks. 

\subsection{Low-Level Search in \ccbsds}

The low-level search in \ccbsds sorts the positive constraints in ascending order of their time dimension and plan to achieve each of them in that order. 
For example, assume there is a single positive constraint $(i, \text{move(A,B)}, [t_i, t_i^u))$.  
Then, the low-level search works by first 
(1) searching for a plan from $s_i$ to $A$ that ends in the time range $[t_i, t_i^u)$, 
then (2) performing the action landmark (i.e., move from $A$ to $B$),
and finally (3) searching for a plan from $B$ to $g_i$ (starting immediately after the action landmark is performed). 

However, in \ccbsds there is an additional challenge for the low-level search: there may be more than one plan to perform each landmark. 
In our example above, there may be an infinite amount of plans from $s_i$ to $A$  that ends in the time range $[t_i, t_i^u)$. 
As we show below, choosing the plan that performs the action landmark earliest does not necessarily lead to finding an optimal solution and might even lead to incompleteness, especially when there are both positive and negative constraints. 

\subsubsection{Example}

\begin{figure}
    \centering
    \includegraphics[width=0.9\columnwidth]{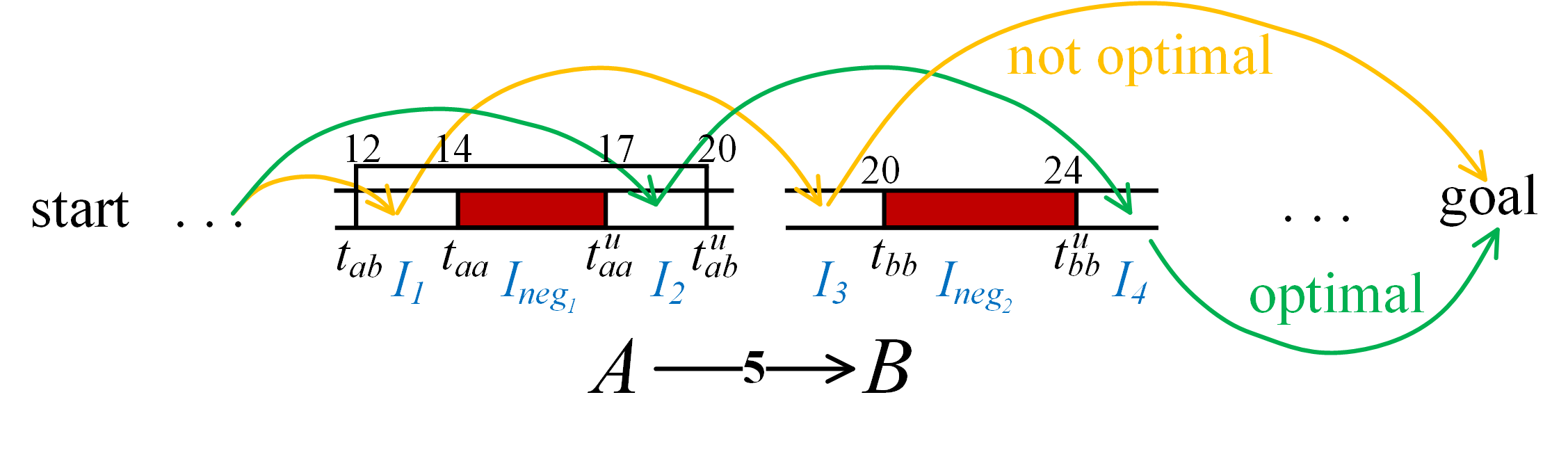}
    \caption{An example where performing the action landmark as early as possible leads to a suboptimal plan.}
    \label{fig:early_plan_not_optimal}
\end{figure}
\begin{figure}
    \centering
    \includegraphics[width=0.9\columnwidth]{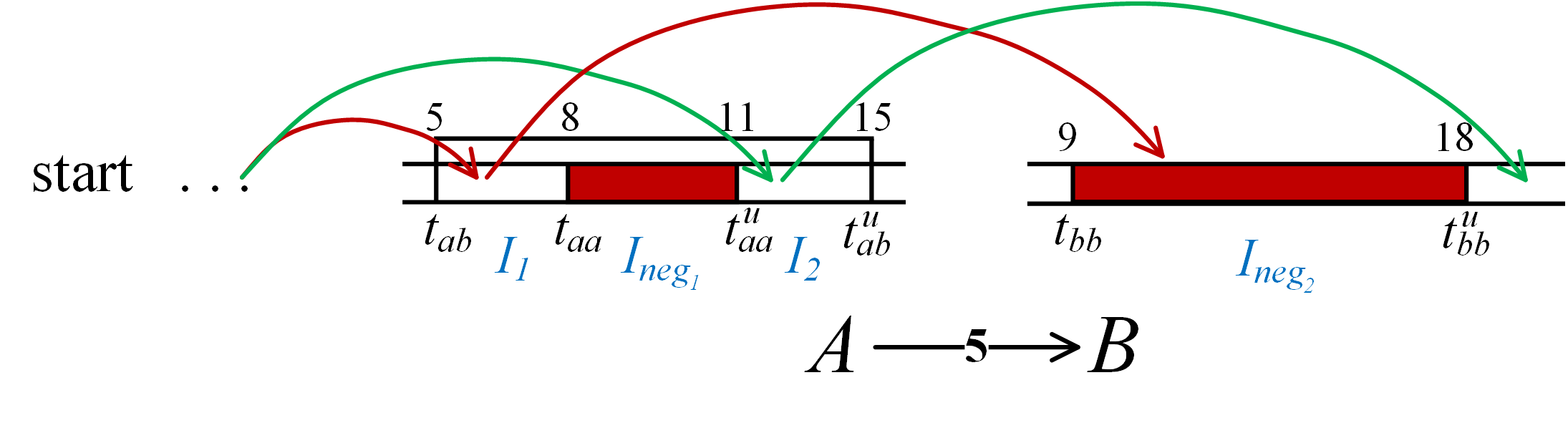}
    \caption{An example where performing the action landmark as early as possible results in failing to find a plan.}
    \label{fig:early_plan_not_complete}
\end{figure}
Consider the illustration depicted in Figure~\ref{fig:early_plan_not_optimal}. 
The low-level search needs to find a plan that satisfies three constraints: 
\begin{itemize}
    \item A positive constraint $(i, \text{move(A,B)}, [t_{ab}, t_{ab}^u)$. 
    \item A negative constraint $(i, \text{wait\_at}(A), [t_{aa}, t_{aa}^u)$.
    \item A negative constraint $(i, \text{wait\_at}(B), [t_{bb}, t_{bb}^u)$.
\end{itemize}
where $t_{ab} < t_{aa} < t_{aa}^u < t_{ab}^u$. 
Thus, the negative constraint on waiting at $A$ ($\text{wait\_at}(A)$) creates two safe intervals for $A$,  $I_1=[0,t_{aa}]$ and $I_2=[t_{aa}^u,\infty)$ that overlap the interval of the positive constraint. The negative constraint on waiting at $B$ ($\text{wait\_at}(B)$) creates two safe intervals for $B$,  
$I_3=[0,t_{bb}]$ and $I_4=[t_{bb}^u,\infty)$. 

Now assume that there are two plans that satisfy the action landmark for the positive constraint, one that reaches $A$ before $t_{aa}$ (shown in yellow) and one that reaches $A$ after $t_{aa}^u$ (show in green). 
Clearly, the lowest-cost plan to achieve the action landmark is the one that reaches $A$ before $t_{aa}$, but to find the optimal solution one must use the second plan. 
Figure~\ref{fig:early_plan_not_complete} illustrates an even more extreme case, where choosing to lowest-cost plan that achieves the action landmark can not be extended to a full plan, because it reaches $B$ during its unsafe interval (marked in red).

\subsubsection{Generalized \sipp}

There may be infinite plans that satisfy an action landmark $l=(i, \text{move}(A,B), [t,t^u))$, i.e., reach $A$ within $[t,t^u)$. Finding only the least-cost plan might lead to incompleteness, as we showed above. To guarantee completeness and optimality we need to find the lowest-cost plan of reaching $A$ \emph{for every safe interval} of $A$ that overlaps with $[t,t^u)$. Only in this case we can deem that every possibility of performing the action landmark has been explored, which preserves completeness. The optimality is preserved due to finding least cost plan of reaching $A$ for every safe interval.  

To this end, we create a generalized version of \sipp (\gsipp) such that: (1) it accepts a set of goal states, one per safe interval of $A$ that overlaps with $[t,t^u)$, and (2) it outputs a set of plans, one per goal state. To each of these plans, we concatenate the action landmark itself $\text{move}(A,B)$. These plans may end in different safe intervals in $B$, which then become distinct start states when searching for a plan to get from $B$ to the next landmark. Thus, \gsipp accepts a set of starts states and a set of goal states and outputs a set of plans, one per goal. 
It works as follows. First, the open list is initialized with 
\emph{all} start states. Then, the search proceeds as in regular \sipp, except that the stop criteria is either when the open list is exhausted or when \emph{all} goal nodes are expanded. The worst case runtime of both \gsipp and \sipp is the same, corresponding to expanding all states in the (vertex, safe interval) state space defined by \sipp.

\subsubsection{Pseudo Code}

\begin{algorithm}[t]
    \KwIn{Negative constraints $C^{(-)}$}
    \KwIn{Positive constraints $C^{(+)}$}
    \KwIn{Agent $i$}
    $\cal S\gets$ ComputeSafeIntervals($C^{(-)}$)\\
    $\cal L \gets$ ComputeLandmarks($C^{(+)}$, $\cal S$)\\
    Starts $\gets \{s_i\}$\\
        \ForEach{landmark $l=(i,\text{move}(A,B),[t,t^u))$ in $\cal L$}{
                Goals $\gets$ computeGoals($l$)\\
                Plans $\gets$ \gsipp(Starts, Goals)\\
                Starts $\gets \emptyset$\\
                \ForEach{plan in Plans}{
                    Append $\text{move}(A,B)$ to $plan$\\
                    Add last state in $plan$ to Starts\\

                }
                Starts $\gets$ Prune Plans/Starts if possible
            }
		\Return \sipp(Starts, $g_i$)
	\caption{Low-level search for \ccbs with \ds} 
\label{alg:low-level-ccbs-ds}
\end{algorithm}

Finally, we can describe the pseudo-code for the \ccbsds low-level search. It accepts a list of negative and positive constraints for an agent $i$. 
Initially, the low-level search computes the safe intervals of every vertex based on the negative constraints (Line 1). 
Then, it computes the action landmarks based on the positive constraints (Line 2). 
These landmarks are sorted by time, and then it iterates over these landmarks (Line 3). 
For each action landmark $l=(i,\text{move}(A,B),[t,t^u))$, it computes the safe intervals of $A$ that intersect with $[t,t^u)$ (Line 5). 
Every such safe interval is considered a goal for \gsipp. 
When all such goals are added, we run \gsipp to find a set of plans, one per goal (Line 6). Then, for each found plan we concatenate the action $\text{move}(A,B)$ to its end (Lines 9-10). If $B$ is not reachable within a safe interval then the plan is discarded. If two or more concatenated plans safely reach $B$ in the same interval $I_{safe_k}$ we prune such plans leaving the only one that reaches this interval earlier (Line 11). A node $(B, I_{safe_k})$ now becomes one of the start nodes for the subsequent search and is added to Starts.

Note that the number of plans satisfying each landmark $l$ is proportional to the number of the negative constraints over the wait actions for the target vertex of $l$. 
Consequently, if no such constraints exist then only one plan to this landmark will be present after pruning (no matter with how many different start and goal nodes the search was initialized). 
In general, in the process of the iterative invocation of the modified \sipp and plan pruning, numerous plans constructed so far might eventually collapse to a single one. This definitely happens when one the planning to the goal is carried out. The reason is that the goal is defined by a single graph vertex and a single time interval ending with $\infty$ as we assume that the agent arrives to its goal and stays there forever. Thus, even if numerous plans to the preceding landmark were found they all will collapse into a single one, i.e. the one that achieves the goal at the earliest possible time which is what \ccbs requires.

\section{Prioritizing Conflicts}




\acf{PC} \cite{BoyarskiFSSTBS15} is the second \cbs enhancement we migrate to \ccbs. 
\pc is a heuristic for choosing which conflict to resolve when expanding a \ct node. 
Different ways to choose conflicts in practice often lead to \ct of different sizes, 
thus have a significant effect on the overall runtime. 
\pc systematically prioritizes conflicts by classifying each conflict as either \emph{cardinal}, \emph{semi-cardinal}, and \emph{non-cardinal}. 
A conflict is called cardinal \emph{iff} splitting a \ct node $N$ over it results in two child nodes
whose cost is higher than the cost of $N$. 
A conflict is semi-cardinal \emph{iff} if the cost of only one child increases while the cost of the other does not. 
A conflict that is not cardinal or semi-cardinal is non-cardinal. 
\cbs with \pc prefers cardinal conflicts to semi-cardinal and semi-cardinal to non-cardinal.  
This way of prioritizing conflicts results in a significant reduction of the expanded \ct nodes compared to vanilla \cbs and makes the algorithm much faster in practice.

In \mapfr, most conflicts are cardinal, i.e., the agents involved in that conflicts are not able to find the paths that respect the corresponding constraints and are of the same cost as before. 
This is because the ability to perform wait actions of arbitrary duration paired with non-uniform move action durations reduces \emph{symmetries}. 
By ``symmetry'' here we mean having multiple shortest paths that have exactly the same cost. 
Thus, differentiating the conflicts based just on their cardinality type is insufficient.

To this end, we propose a generalized version of \pc that introduces a finer-grained prioritization of conflicts, by introducing the notion of \emph{cost impact}. 
Intuitively, the cost impact of a conflict is how much the cost of the solution is increased when it is resolved. More formally, for a \ct node $N$ with a \ccbs conflict $Con=(a_i, t_i, a_j, t_j)$, 
let $N_i$ and $N_j$ be the \ccbs nodes obtained by splitting over this conflict, 
and let $\delta_i$ be the difference between the cost of $N$ and $N_i$. 
We define the \emph{cost impact} of the conflict $Con$, denoted $\Delta(Con)$, as $\min(\delta_i, \delta_j)$ \footnote{We also experimented with $\Delta(Con) = \max(\delta_i, \delta_j)$ and $\Delta(Con) = \sum(\delta_i, \delta_j)$ but the affect on performance was minimal.}.
Our adaptation of \pc to \ccbs chooses to split a \ct node on the conflict with the largest cost impact. This follows the same rationale as \pc, as we prioritize the resolution of conflicts that will
reveal the highest unavoidable cost that was so far hidden in conflicts.

\section{Heuristics for High-Level Search}

To guarantee optimality, the high-level search in \cbs explores the \ct tree in a best-first fashion. \citet{felner2018adding} and \citet{CBSH2} introduced admissible heuristics to the \cbs high-level search. 
These heuristics estimate the difference in cost between a CT node and the optimal solution. Both heuristics are \emph{admissible}, i.e., they are a lower bound on the actual cost difference, and therefore can be safely added to the cost of a \ct node when choosing which node to expand next. 
Indeed, these heuristics were shown to significantly decrease the number of the expanded \ct nodes and improve the performance of \cbs.

Drawing from these works we suggest two admissible heuristics for \ccbs. 
The first admissible heuristic, denoted $H1$, is based on solving the following linear programming problem (LPP). 
This LPP has $n$ non-negative variables $x_1,\ldots x_n$, one for each agent. 
Each conflict $Con_{i,j}$ between agents $i$ and $j$ 
in the \ct node for which we are computing the heuristic
introduces the LPP constraint $x_i + x_j \geq \Delta(Con_{i,j})$. 
The objective to be minimized is $\sum_{i=1}^n{x_i}$. 
By construction, for any solution to this LPP, the value $\sum_{i=1}^n{x_i}$ is an admissible heuristic since for every conflict $Con_{i,j}$ the solution cost is increased by at least $\Delta(Con_{i,j})$.

The second admissible heuristic we propose, denoted $H2$, follows $h_1$ the approach suggested in~\cite{felner2018adding}. 
There, the heuristic was based on identifying \emph{disjoint cardinal conflicts}, which are cardinal conflicts between disjoint pairs of agents. 
As discussed above, in \ccbs most conflicts are cardinal but their \emph{cost impact} can vary greatly. 
Therefore, in the $H2$ heuristic we aim to choose the disjoint cardinal conflicts that would have the largest cost impact. We do so in a greedy manner, sorting the conflicts in $N.\Pi$ in descending order of their cost impact. Then, conflicts are picked one by one in this order. 
After a conflict is picked, we remove from the conflict list all conflicts that involve any of the agents in this conflict. This continues until all the conflicts are either picked or removed. 
The $H2$ heuristic is the sum of the cost impacts of the chosen conflicts. 
By construction the chosen conflicts are disjoint and so $H2$ is admissible. 
While $H2$ is less informed than $H1$ (the one computed by solving LPP), it is faster to compute. 
We observed experimentally that the practical difference between these heuristics was negligible -- an average difference of 1\%. We conjecture that the reason $H1$ and $H2$ perform similarly is that often the conflict graph consists of disjoint pairs of connected agents, in which case the minimum vertex cover ($H1$) would also be found by the simple greedy approach ($H2$).
In our experiments described below we used $H2$ and refer to it as H. 

\section{Empirical Evaluation}

\begin{figure*}[t]
    \centering
    \includegraphics[width=\textwidth]{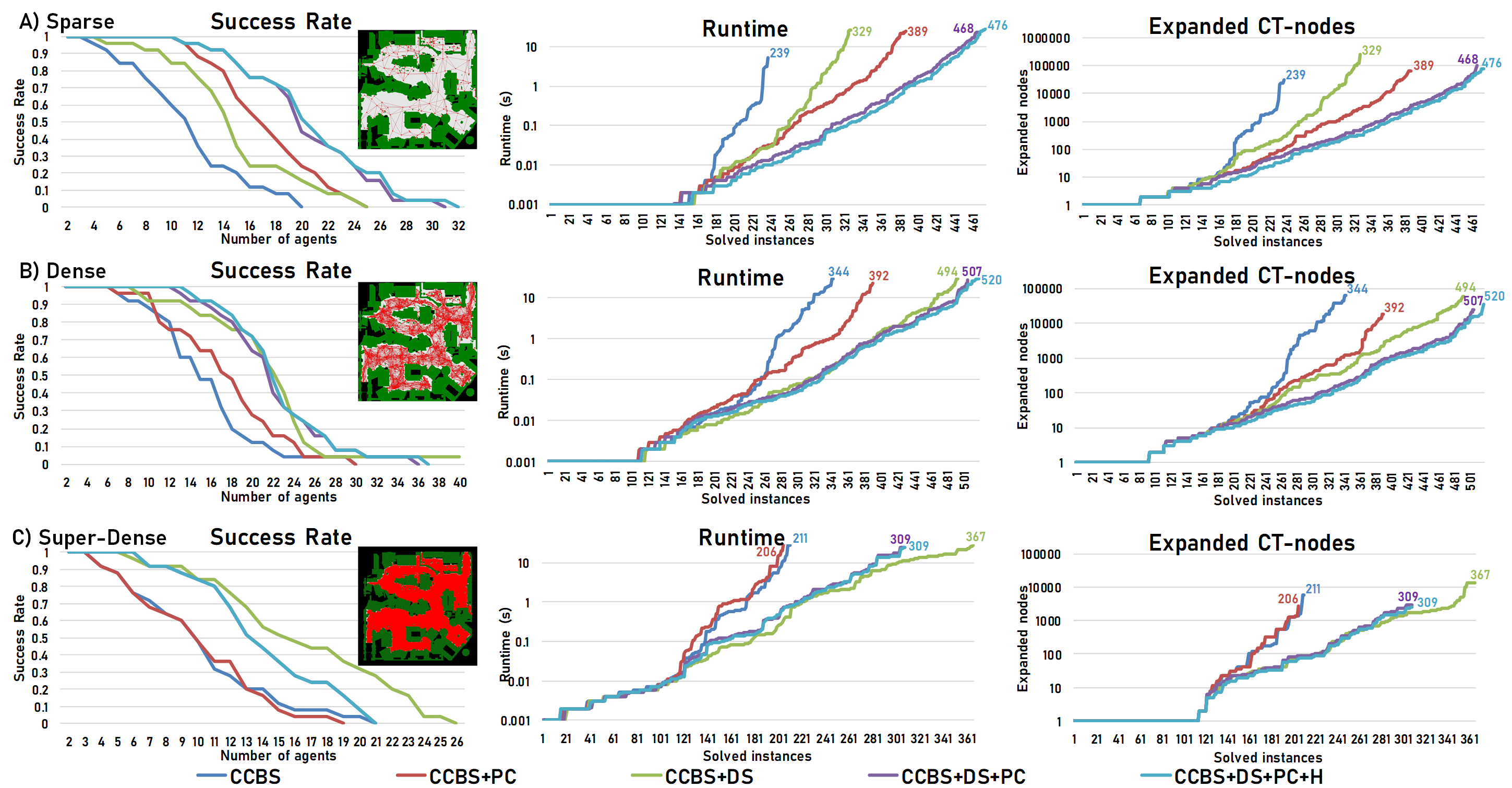}
    \caption{The performance of \ccbs and its variants on the sparse, dense and super-dense roadmaps.}
    \vspace{-0.3cm}
    \label{figRoadmapsResults}
\end{figure*}

\begin{figure*}[t]
    \centering
    \includegraphics[width=\textwidth]{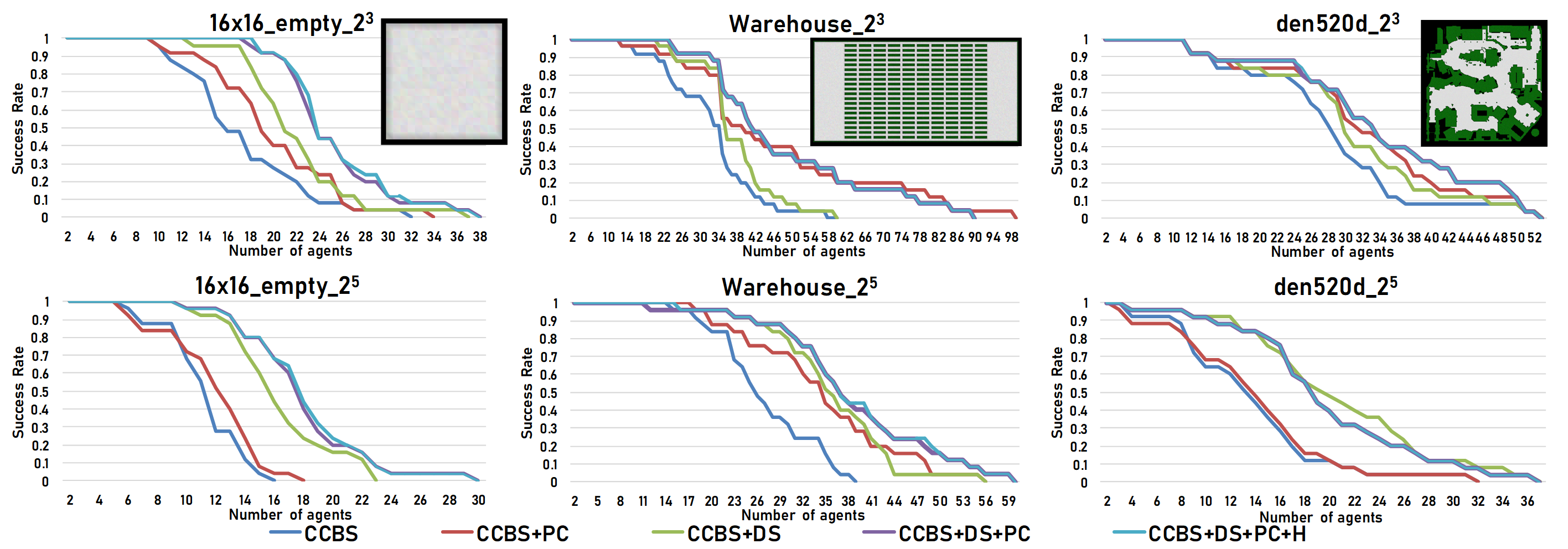}
    \caption{Success rates for \ccbs and its modifications on different $2^k$-connected grids.}
    \label{figGridsResults}
\end{figure*}

We have incorporated all the \ccbs enhancements described so far and evaluated different versions of \ccbs in different \mapfr scenarios involving general graphs (roadmaps) and grids.\footnote{Our implementation and all the raw results are available at: \texttt{github.com/PathPlanning/Continuous-CBS}.} 
Specifically, we evaluated the basic \ccbs, \ccbs with \pc (\ccbs+\pc), \ccbs with \ds (\ccbsds), 
\ccbs with both \ds and \pc (\ccbs+\ds+\pc), 
and \ccbs with all the improvements (\ccbs+\ds+\pc + H). 
In the conducted experiments all agents were assumed to be disk-shaped with radius equal to $\sqrt{2}/4$.

In each run of the evaluated algorithm, we recorded the runtime, the number of expanded \ct nodes, and whether the algorithm was able to find a solution under a time limit of 30 seconds. We chose this specific limit to demonstrate near real time performance. Moreover, in the preliminary experiments with different time limits (from 1s to 300s) we observed that the difference in performance of \ccbs with 30s time limit and 300s time limit is not significant. 

\subsection{Implementation Details}

Conflict detection in \mapfr is more involved than in classical \mapf and is more computationally intensive. 
To compensate for that we have implemented the following approach to cache the intermediate conflict detection results and speed up the search. 
We detect all the conflicts in the root \ct node and store them with the node. 
After choosing a conflict and performing a split we copy all the conflicts to a successor node except the ones involving the agent that has to re-plan its path. 
After such re-planning, newly introduced conflicts (if any) are added to the set of conflicts for that \ct node. Indeed, this leads to a memory overhead, which in our experiments varied from 15\% to 250\%, depending on how many conflicts were discovered. 

To compute the \emph{cost impacts} of the conflicts for versions of \ccbs that use \pc or the high-level search heuristic $H$, we run the low-level search explicitly to resolve these conflicts and acquire the needed cost increase values.

To speed-up the low-level search, we pre-compute a set of heuristics, $h_1$, ..., $h_n$ to estimate cost-to-go to each goal. 
To compute $h_i$ we run Dijkstra's algorithm with $g_i$ as the source node. Such heuristics are more informative compared to Euclidean distance but their computation complexity is polynomial in the graph size. However, the runtime needed to compute all heuristics is significantly less than overall runtime of solving the \mapfr problem.
When \ds is used, the low-level search performs multiple searches to achieve the landmarks created by the positive constraints. When searching for the intermediate goals associated with each landmarks, we implemented a Differential Heuristic (DH)~\cite{goldenberg2011compressed} with the pre-computed heuristics $h_1,\ldots, h_n$ as pivots.

\subsection{Evaluation on the Roadmaps}

In the first set of experiments we have evaluated \ccbs on 3 different roadmaps, referred to here as \emph{sparse}, \emph{dense} and \emph{super-dense}. The sparse roadmap contains 158 nodes and 349 edges, the dense roadmap contains 878 nodes and 7,341 edges, and the super-dense roadmap contains 11,342 vertices and 263,533 edges. All of these graphs were automatically generated by applying a roadmap-generation tool from the Open Motion Planning Library (OMPL)~\cite{sucan2012ompl} on the \texttt{den520d} map from the game Dragon Age Origin (DAO). This map is publicly available in the MovingAI \mapf benchmark~\cite{stern2019multi}. 

For each roadmap, 25 different scenarios were generated. Each scenario is a list of start-goal vertices, chosen randomly from the graph. 
Then, we pick the first $n=2$ start-goal pairs and create a \mapfr instance for $n$ agents. If the evaluated algorithm solves this instance within the 30 seconds time limit, we proceed by increasing $n$ by 1 and creating a new \mapfr instance. 
This is repeated until the evaluated algorithm is not able to solve the instance in 30 seconds. We then proceed to the next scenario. 

The results are shown in Fig.\ref{figRoadmapsResults}. Consider first the success rate plots (left). The first clear trend we observe is that all the proposed \ccbs improvements are significantly better than the baseline \cbs in almost all cases. 
E.g., on the dense roadmap \ccbs+\ds+\pc+H manages to achieve 0.8 success rate for the instances with 20 agents, while \ccbs success rate for this number of agents is only 0.1.

Next, consider the relative performance of \ccbs with different combinations of improvements. In general, the most advanced version of the algorithm, i.e. \ccbs+\ds+\pc+H, outperforms the competitors on sparse and dense roadmaps. 
However on the super-dense this is not the case. On this roadmap, \ccbs+\ds+\pc+H is dominated by \ccbs+\ds which was able to solve 25 agents while the former -- 20. 
Indeed, in this roadmap the \pc component on its own is ineffective, as can be seen when comparing the basic \ccbs and \ccbs+\pc. 
We explain this behavior by observing that this roadmap has a very high branching factor (every vertex has almost 50 neighbors on average). 
This helps to eliminate conflicts by finding an appropriate detour of nearly the same cost. Thus the cost impacts, which are computationally intensive to compute, are very low and provide limited value in differentiating between the conflicts. 

Next, consider the runtime and expanded \ct nodes plots in Figure~\ref{figRoadmapsResults}. 
These plots are built in the following fashion. 
Each data point $(x,y)$ on a plot says that an algorithm was able to solve $x$ problem instances within $y$ seconds/\ct nodes expansions. 
For example, on the dense roadmap \ccbs solved only 276 instances in less than 1 second, \ccbs+PC -- 340 instances, while \ccbs+\ds+\pc+H -- 404. In general, the closer the line to $x$-axis and the longer it is -- the better. The values at the end of the lines show the exact numbers of the solved instances. 

The general trend for runtime and high-level expansions are similar to the ones for the success rate: 
\ccbs+\ds+\pc+H is the best on \emph{sparse} and \emph{dense} roadmaps and \ccbs+\ds is the best on \emph{super-dense}. These results highlight our improvement over vanilla \cbs, where our best \ccbs version is up to 2 orders of magnitude faster in some cases.

We also analyzed separately the impact of adding the high-level heuristic ($H$) on the instances that involve large numbers of \ct expansions. We took the results of 100 instances with the highest values of expanded \ct nodes solved by \ccbs+DS+PS and \ccbs+DS+PC+H averaged the number of expansions and compared them. The number of expansions for \ccbs+DS+PC+H was lower by 26.5\%, 21.6\% and 17.8\% for sparse, dense and super-dense roadmaps respectively. Thus, adding heuristic proved to be a valuable technique, especially for the hard instances involving large number of expansions.

\subsection{Evaluation on Grids}

The second set of experiments we conducted was on 8-connected ($2^3$) and 32-connected ($2^5$) grids from the MovingAI \mapf benchmark~\cite{stern2019multi}. We used a 16x16 empty grid (16x16\_empty), a warehouse-like grid (warehouse-10-20-10-2-2), and a grid representation of the den520d DAO map. 
Here we used the 25 scenario-files supplied by the \mapf benchmark for each grid. 
The results of the second series of experiments are shown in Fig.\ref{figGridsResults}.

Here we can see that in almost all cases  
the best results were obtained by \ccbs with all our enhancements (\ccbs+\ds+\pc+H). 
Comparing the results on grids with different connectedness, one can notice the same trend as observed for roadmaps with respect to the benefit of \pc and \ds: increasing the branching factor makes \pc less effective and \ds more effective.
This benefit for \ds is explained by the fact that positive constraints help to reduce the branching factor by reducing the amount of possible alternative trajectories to one. Thus, higher branching factor means stronger pruning by positive constraints.

\begin{table}
\resizebox{\columnwidth}{!}{
\begin{tabular}{| c | c c | c c | c c | c c |}
\hline
 & \multicolumn{2}{|c|}{PC} & \multicolumn{2}{c|}{DS}& \multicolumn{2}{c|}{DS+PC}& \multicolumn{2}{c|}{DS+PC+H}\\
& k=3 & k=5 & k=3 & k=5 & k=3 & k=5 & k=3 & k=5\\
\hline
16x16 & 33.10\% & 72.15\% & 13.97\% & 14.85\% & 6.72\% & 10.25\% & 5.59\% & 9.77\% \\
warehouse & 14.04\% & 15.69\% & 28.64\% & 23.70\% & 10.84\% & 18.36\% & 10.78\% & 14.31\% \\
den520d & 31.25\% & 100.00\% & 37.50\% & 67.71\% & 17.42\% & 76.42\% & 14.29\% & 67.11\% \\
\hline
\end{tabular}
}
\caption{The ratio of expanded CT-nodes between \ccbs and its modifications on grids (lower = better).}
\label{TableGridsExpansions}
\end{table}

Finally, we considered the 100 instances in each grid for which basic \ccbs expanded the most \ct nodes. 
Table~\ref{TableGridsExpansions} presents the median of the ratios of expansions between basic \ccbs and the other versions. 
As one can see, CCBS+DS+PC+H expands the fewest \ct nodes. Also, we observe that in most cases additional connectivity of the grid makes all the enhancements less beneficial. 

\section{Conclusions and Future Work}
In this work, we have proposed three improvements to \ccbs, an algorithm for finding optimal solutions to \mapfr problems in which time is continuous. 
The first \ccbs improvement we proposed, called \ds, changes how \ct nodes are expanded by introducing positive and negative constraints. 
To implement this improvement, we modified the \ccbs low-level search and applied a generalized version of \sipp with multiple start and goal nodes. 
The second improvement, called \pc, prioritizes the conflicts to resolve by computing the cost of the solution that resolves them. 
The third \ccbs improvement we proposed is two admissible heuristics for the high-level search. 
In a comprehensive experimental evaluation, we observed that using these improvements, \ccbs can scale to solve much more problems than the basic \ccbs, solving in some cases almost twice as many agents. 
Allowing \ccbs to scale to larger problem is key to applying it to a wider range of real-world applications and also as a foundation for more generate \mapf settings in which the underlying graph is also changing rapidly. 

\subsection*{Acknowledgments}
The research for this project is partially funded by ISF grant \#210/17 to Roni Stern, by the RSF grant 
\#20-71-10116 to Konstantin Yakovlev (studies of the generalized SIPP algorithm), and by the RUDN University Strategic Academic Leadership Program to Anton Andreychuk. 

\bibliography{mapf.bib}

\end{document}